\documentclass[conference]{IEEEtran}
\IEEEoverridecommandlockouts
\usepackage{cite}
\usepackage{amsmath,amssymb,amsfonts}
\usepackage{algorithmic}
\usepackage{graphicx}
\usepackage{subfig}
\usepackage{textcomp}
\usepackage{xcolor}
\def\BibTeX{{\rm B\kern-.05em{\sc i\kern-.025em b}\kern-.08em
    T\kern-.1667em\lower.7ex\hbox{E}\kern-.125emX}}

\usepackage{fancyhdr}
\begin{document}

\title{Comparing an android head with its digital twin regarding the dynamic expression of emotions\\
}


\author{\IEEEauthorblockN{1\textsuperscript{st} Amelie Kassner
\IEEEauthorblockA{\textit{Computer Science and Media} \\
\textit{Stuttgart Media University}\\
Stuttgart, Germany \\
ak305@hdm-stuttgart.de}
\and
\IEEEauthorblockN{2\textsuperscript{nd} Christian Becker-Asano}
\IEEEauthorblockA{\textit{Computer Science and Media} \\
\textit{Stuttgart Media University}\\
Stuttgart, Germany\\
becker-asano@hdm-stuttgart.de}
}}

\maketitle
\thispagestyle{fancy}

\begin{abstract}
Emotions, which are an important component of social interaction, can be studied with the help of android robots and their appearance, which is as similar to humans as possible. The production and customization of android robots is expensive and time-consuming, so it may be practical to use a digital replica. In order to investigate whether there are any perceptual differences in terms of emotions based on the difference in appearance, a robot head was digitally replicated. In an experiment, the basic emotions evaluated in a preliminary study were compared in three conditions and then statistically analyzed. It was found that apart from fear, all emotions were recognized on the real robot head. The digital head with "ideal" emotions performed better than the real head apart from the anger representation, which offers optimization potential for the real head. Contrary to expectations, significant differences between the real and the replicated head with the same emotions could only be found in the representation of surprise.
\end{abstract}

\begin{IEEEkeywords}
facial expression, emotion, empirical study, android robot, social robotics
\end{IEEEkeywords}

\section{Introduction and motivation}
Robots are expected to support us as social partners in the future. For natural interaction, it is necessary that their appearance and behavior are adapted to their environment \cite{b1}. Emotions serve as a nonverbal tool that can enhance such interactions \cite{b2}. Android robots can be used for human-human and human-robot interaction research due to their very human-like appearance. However, their expressiveness is limited by their hardware and their production is still quite expensive. Virtual robot heads, on the other hand, can be produced comparatively easy without high cost. A virtual robot head can be used to mimic the human face and better understand its functionality \cite{b3}. Furthermore, virtual robot heads have more freedom of movement in their animation since they are not bound by physical constraints.

In order to investigate possible differences regarding emotion perception between a real and virtual robot head, a physical android robot head was digitally recreated in Unreal Engine 5 (UE5). This replica serves as a basis for further research and provides information about possible adaptations of the real robot head to better represent emotions on it. Six basic emotions \cite{b11} were modelled with the robot head and validated in a pre-experiment. Afterwards, the real and the virtual version of the robot head were compared against each other using a 3D visualization inside a head-mounted display for the virtual version. 

The remainder of this paper is structured as follow. In the following section related work will be presented and discussed. In Section~\ref{sec-ef} the experimental hypotheses will be stated and the hardware setup will be introduced, before a pre-study is explained in Section~\ref{sec-ps}. The main study is described in Section~\ref{sec-ms} with its results presented and analyzed in Section~\ref{sec-results}. A general discussion in Section~\ref{sec-discussion} concludes our presentation.

\section{Related work}

\subsection{Emotions}

There are many different approaches trying to define emotions \cite{b5, b6}, but a unified definition has not been found yet. Emotion theories and models such as basic emotion theory (BET) \cite{b8}, main emotion systems \cite{b9} or prototypical approaches \cite{b10} try to look at emotions from different perspectives. Ekman's research has defined the basic emotions of anger, disgust, fear, happiness, sadness, and surprise, and assumes that they are universal and culturally independent \cite{b11}. Even though the validity of this research has been doubted in some cases (cf.~\cite{b12,b8}), these basic emotions serve as a basis for research in many cases, including in the field of human-robot interaction \cite{b13}. Ekman's results could be investigated and replicated in further studies \cite{b14}. 

\subsection{Social robots and androids}

Social robots are able to interact naturally with humans via verbal and nonverbal signals. Emotions are a part of this and can help represent a robot's internal state and allow viewing individuals to interpret and respond to it \cite{b15}. Androids have an appearance as similar to humans as possible and are intended to advance research regarding human-human and human-robot interaction. Human movements and facial expressions are of great importance here for natural interaction \cite{b16}. Geminoid HI-1 and Geminoid F have already been used to study cross-cultural differences in terms of emotion perception, where fear was more difficult to detect and confusion varied depending on nationality \cite{b2}. Replicating human faces is difficult due to their high complexity \cite{b3}, yet this has been attempted several times. Robot heads are often able to represent emotions, which has been validated in experiments over images or the real robots \cite{b17, b18}. Since representation via two-dimensional displays can be problematic, a hybrid approach, for example, relies on a robot head with facial expressions controlled via computer animation \cite{b3}.

\subsection{Virtual Agents}

Virtual agents with an embodied representation need emotions as they can be an important part of their credibility, higher satisfaction, better performance, and better perception of the agent \cite{b19, b20}. Virtual agents have already been used to explore whether age can have an impact on emotion recognition \cite{b21}. Creating a virtual agent is complex and requires a multidisciplinary approach for correct processing of signals, appropriate animations, and underlying knowledge, for example on emotion representation \cite{b22}. 

\subsection{Differences between real and virtual representations}

When interacting with real and virtual robots, presence and embodiment play an important role and must be considered when making comparisons \cite{b3, b24}. When represented via screens, virtual robots are limited in their interaction capabilities \cite{b25}. Interactions with physical robots are also often perceived more positively \cite{b15, b26, b28, b29}. Instead of using a representation on a screen, it is possible to use virtual reality (VR) and resort to representations via CAVE \cite{b30} or head-mounted displays (HMD). The depicted robot and the person viewing it are then co-present \cite{b31}. Using the same experimental setup in real and virtual allows a direct comparison as long as differences of the media are taken into account \cite{b32}. Contrary to previous results, however, no significant differences between the presentation in real and in VR could be found in \cite{b33}.

\section{Experiment foundations}
\label{sec-ef}

\subsection{Hypotheses}

The research is based on the following hypotheses:

\begin{enumerate}
  \item There are general differences in emotion perception between the real robot head, the virtual version of the robot head with recreated emotions, and the virtual version of the robot head with emotions that are not artificially limited to recreate the physical constraints of the real robot head.
  \item On the virtual robot head, the emotions without constraints will be evaluated better than the emotions modeled on the real robot head.
  \item The real robot head will be rated better than the digital robot head with the recreated emotions.
  \item Similar to the second and third hypothesis, the virtual robot head with unrestricted emotion representation will be evaluated better than the real robot head. 
\end{enumerate}

The last hypothesis relies on the assumption that the better emotion representation achieved by more freedom of animation has a greater influence than the different types of embodiment (real/virtual).

\subsection{Hardware and software used}
An android robot head of the Stuttgart Media University was used for both the main and pre-study. This head is equipped with 14 actuators powered by compressed air for controlling facial expressions and head movement.

For the digital replica, the robot head was scanned with the EinScan HX handheld 3D scanner, post-processed in Blender, and rigged and animated via the UE5's MetaHuman framework. The virtual scene was rendered using the HTC Vive connected to a laptop equipped with Intel i7-10870H CPU, 32 GB RAM and NVIDIA GeForce RTX 3080 GPU.

\section{Pre-study}
\label{sec-ps}
In order to validate the emotions, a pre-study was conducted in the form of an online survey. The experimental setup of \cite{b2} was used as a reference. The emotions anger, sadness, surprise, fear, happiness, disgust and a neutral facial expression were implemented on the real robot head (condition 1). Those emotions were manually recreated on the virtual robot head using the MetaHuman facial rig (condition 2). Additionally, another set of emotional expressions were taken from the MetaHuman pose library (condition 3), thus, ignoring the hardware limitations of the real robot head.  

Participants were randomly assigned to one of the three conditions. At the beginning, participants had to read through and confirm a privacy statement. Age, gender, highest level of education, affinity for technology, and previous experience with android robots were queried. Afterwards, it was explained that labels should be assigned to each of the images of emotional facial expressions. It was pointed out that the option "None of these labels" can be selected should none of the listed labels fit, and that it is possible to assign labels more than once. On the next page, images representing neutral, anger, sadness, surprise, fear, happiness, and disgust were displayed below each other in random order. All labels available for selection were introduced at the beginning of the page. Participants were asked to view all images first, before assigning labels.

From March 4 to March 10, 2023, participants were invited to take part in the online survey through the University's mailing list and social networks. A total of 127 people participated, 114 of whom completed the survey and were included. 70 individuals identified themselves as female ($M = 31.06$ years; $SD = 15.35$), 41 as male ($M = 29.95$ years; $SD = 12.81$), and 3 as non-binary ($M = 26$ years; $SD = 3.61$). Of these, 37 subjects had been automatically assigned to the first condition (real robot head), 30 to the second condition (virtual head with non-constraint emotions), and 47 to the third condition (virtual head with replicated emotions). The results are presented in Tables \ref{tab_con1}, \ref{tab_con2} and~\ref{tab_con3}.  

In the first condition, happiness (91.9\%), sadness (89.2\%), and neutral facial expression (89.2\%) were well recognized. Surprise was frequently mistaken for fear (21.6\%), and anger for disgust (27\%). Fear (10.8\%, below chance level) and disgust (18.9\%) were both poorly recognized, with disgust most often classified as none of the emotions (51.4\%) and fear as surprise (73\%).

\begin{table}[htbp]
\caption{Confusion matrix for emotion recognition of the first condition (real robot head) of the pre-study, data in percent; rows show selected label, columns actual depicted emotion}
\begin{center}
\begin{tabular}{|c|c|c|c|c|c|c|c|}
\hline
\multicolumn{8}{|c|}{\textbf{Condition 1 (N=37)}} \\
\hline
&\multicolumn{7}{|c|}{\textbf{Picture}} \\
\hline
\textbf{Label} & \textbf{\textit{neut.}}& \textbf{\textit{angr.}}& \textbf{\textit{sad}}& \textbf{\textit{surp.}}& \textbf{\textit{fear}}& \textbf{\textit{happ.}}&\textbf{\textit{disg.}} \\
 \hline
 neutral& \textbf{89.2$^{\mathrm{*}}$}& 2.7& 2.7& 2.7& 2.7& 5.4& 16.2$^{\mathrm{*}}$\\
\hline
 anger& 0& \textbf{48.6$^{\mathrm{*}}$}& 0& 8.1& 0& 0& 0\\
\hline
 sadness& 0& 0& \textbf{89.2$^{\mathrm{*}}$}& 0& 2.7& 0& 0\\
\hline
 surprise& 0& 0& 0& \textbf{64.9$^{\mathrm{*}}$}& \textbf{73$^{\mathrm{*}}$}& 0& 5.4\\
\hline
 fear& 0& 0& 2.7& 21.6$^{\mathrm{*}}$& 10.8& 0& 0\\
\hline
 happiness& 5.4& 0& 2.7& 2.7& 0& \textbf{91.9$^{\mathrm{*}}$}& 8.1\\
\hline
 disgust& 0& 27& 0& 0& 0& 0& 18.9$^{\mathrm{*}}$\\
\hline
 none& 5.4& 21.6$^{\mathrm{*}}$& 2.7& 0& 10.8& 2.7& \textbf{51.4$^{\mathrm{*}}$}\\
\hline
\multicolumn{8}{r}{$^{\mathrm{*}}$above chance level of 12.5\%}
\end{tabular}
\label{tab_con1}
\end{center}
\end{table}

In the second condition, very good recognition rates were achieved for sadness (96.7\%), surprise (96.7\%), fear (90\%), and happiness (100\%). Anger, on the other hand, was more often classified as disgust (46.7\%) and vice versa for disgust (anger 43.3\%).

\begin{table}[htbp]
\caption{Confusion matrix for emotion recognition of the second condition (virtual robot head with "ideal" emotions) of the pre-study, data in percent; rows show selected label, columns actual depicted emotion}
\begin{center}
\begin{tabular}{|c|c|c|c|c|c|c|c|}
\hline
\multicolumn{8}{|c|}{\textbf{Condition 2 (N=30)}} \\
\hline
&\multicolumn{7}{|c|}{\textbf{Picture}} \\
\hline
\textbf{Label} & \textbf{\textit{neut.}}& \textbf{\textit{angr.}}& \textbf{\textit{sad}}& \textbf{\textit{surp.}}& \textbf{\textit{fear}}& \textbf{\textit{happ.}}&\textbf{\textit{disg.}} \\
 \hline
 neutral& \textbf{100$^{\mathrm{*}}$}& 0& 0& 0& 0& 0& 0\\
\hline
 anger& 0& 40$^{\mathrm{*}}$& 0& 8.1& 0& 0& \textbf{43.3$^{\mathrm{*}}$}\\
\hline
 sadness& 0& 0& \textbf{96.7$^{\mathrm{*}}$}& 0& 3.3& 0& 6.7\\
\hline
 surprise& 0& 0& 0& \textbf{96.7$^{\mathrm{*}}$}& 6.7& 0& 0\\
\hline
 fear& 0& 0& 0& 0& \textbf{90$^{\mathrm{*}}$}& 0& 3.3\\
\hline
 happiness& 0& 0& 0& 3.3& 0& \textbf{100$^{\mathrm{*}}$}& 0\\
\hline
 disgust& 0& \textbf{46.7$^{\mathrm{*}}$}& 0& 0& 0& 0& 20$^{\mathrm{*}}$\\
\hline
 none& 0& 13.3$^{\mathrm{*}}$& 3.3& 0& 0& 0& 26.7$^{\mathrm{*}}$\\
\hline
\multicolumn{8}{r}{$^{\mathrm{*}}$above chance level of 12.5\%}
\end{tabular}
\label{tab_con2}
\end{center}
\end{table}

Regarding the third condition, sadness ($91.5\%$), surprise, and happiness ($87.2\%$ each) performed best. It is interesting that neutral facial expression, although corresponding to the same facial expression as in condition two, performed worse here ($68.1\%$). As in condition one, anger was mistaken for disgust ($12.8\%$). Fear was also below chance ($8.5\%$) and was classified significantly more often than surprise ($57.4\%$). Disgust was more likely to be classified as "None of these labels" ($38.3\%$).

\begin{table}[htbp]
\caption{Confusion matrix for emotion recognition of the third condition (virtual robot head with replicated emotions) of the pre-study, data in percent; rows show selected label, columns actual depicted emotion}
\begin{center}
\begin{tabular}{|c|c|c|c|c|c|c|c|}
\hline
\multicolumn{8}{|c|}{\textbf{Condition 3 (N=47)}} \\
\hline
&\multicolumn{7}{|c|}{\textbf{Picture}} \\
\hline
\textbf{Label} & \textbf{\textit{neut.}}& \textbf{\textit{angr.}}& \textbf{\textit{sad}}& \textbf{\textit{surp.}}& \textbf{\textit{fear}}& \textbf{\textit{happ.}}&\textbf{\textit{disg.}} \\
 \hline
 neutral& \textbf{68.1$^{\mathrm{*}}$}& 4.3& 0& 0& 4.3& 10.6& 10.6\\
\hline
 anger& 10.6& \textbf{59.6$^{\mathrm{*}}$}& 0& 0& 0& 0& 19.1$^{\mathrm{*}}$\\
\hline
 sadness& 10.6& 6.4& \textbf{91.5$^{\mathrm{*}}$}& 0& 4.3& 0& 6.4\\
\hline
 surprise& 2.10& 0& 0& \textbf{87.2$^{\mathrm{*}}$}& \textbf{57.4$^{\mathrm{*}}$}& 0& 0\\
\hline
 fear& 4.3& 0& 6.4& 10.6& 8.5& 0& 2.1\\
\hline
 happiness& 0& 0& 0& 0& 4.3& \textbf{87.2$^{\mathrm{*}}$}& 0\\
\hline
 disgust& 2.1& 12.8$^{\mathrm{*}}$& 0& 0& 2.1& 0& 23.4$^{\mathrm{*}}$\\
\hline
 none& 2.1& 17$^{\mathrm{*}}$& 2.1& 2.1& 19.1$^{\mathrm{*}}$& 2.1& \textbf{38.3$^{\mathrm{*}}$}\\
\hline
\multicolumn{8}{r}{$^{\mathrm{*}}$above chance level of 12.5\%}
\end{tabular}
\label{tab_con3}
\end{center}
\end{table}

\section{Main study}
\label{sec-ms}
Based on the results of the pre-study, disgust was excluded for the main study due to the poor recognition rate in all three conditions. Since fear was more often classified as surprise, the representation of fear was modified to be more distinguishable from surprise. The main experiment followed a within-subject design. The order of the conditions was determined via the Balanced Latin Square Design and the sequence of emotional expressions shown within each condition was randomized.

\subsection{Displayed emotions and their evaluation}
In the main study, the same three conditions as in the pre-study were used, with the difference of showing live transitions of the emotions from the neutral expression into the emotion. The timings for the animations of the different expressions were taken from \cite{b13} with, e.g., surprise being presented much faster than happiness. The virtual environment was built to look as similar as possible to the actual environment.

\begin{figure*}[t]
\centering
\begin{tabular}{c@{}c@{}c@{}c@{}c@{}c@{}}
\subfloat{\includegraphics[scale=0.1]{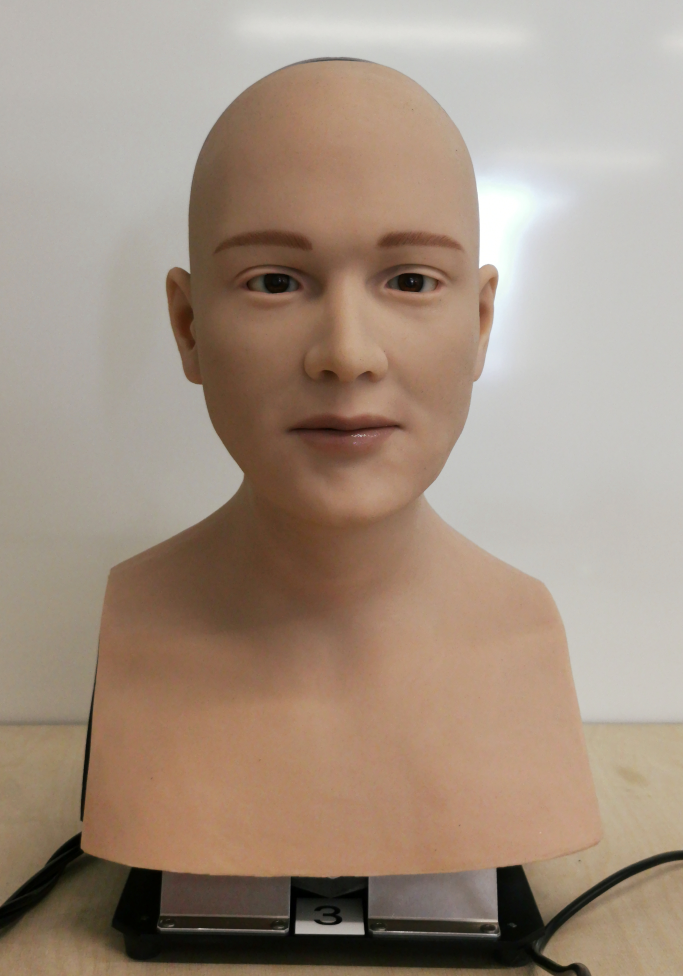}} &
\subfloat{\includegraphics[scale=0.1]{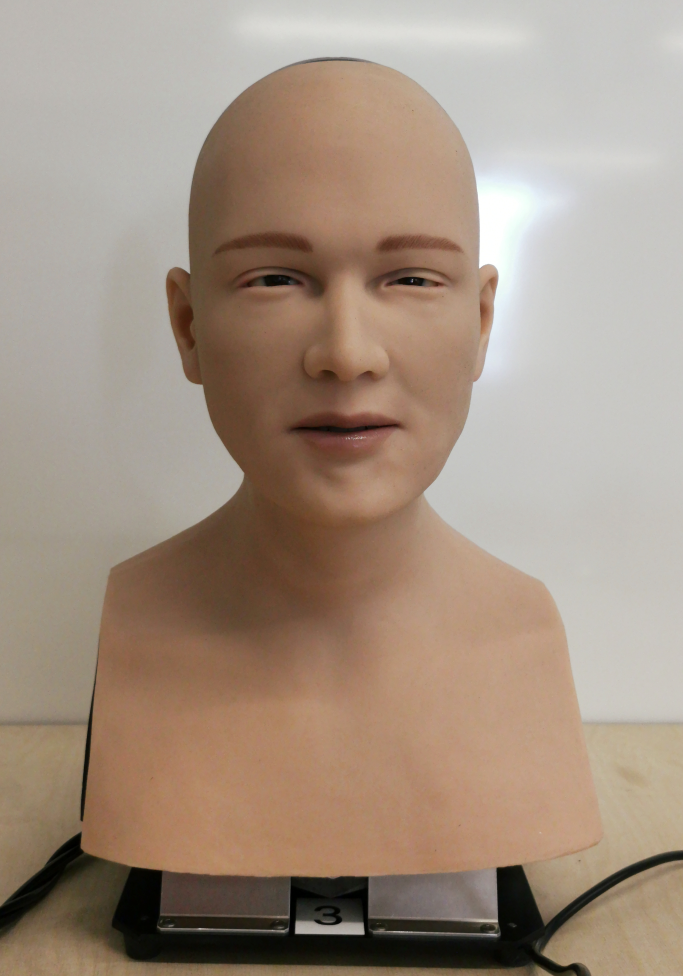}} &
\subfloat{\includegraphics[scale=0.1]{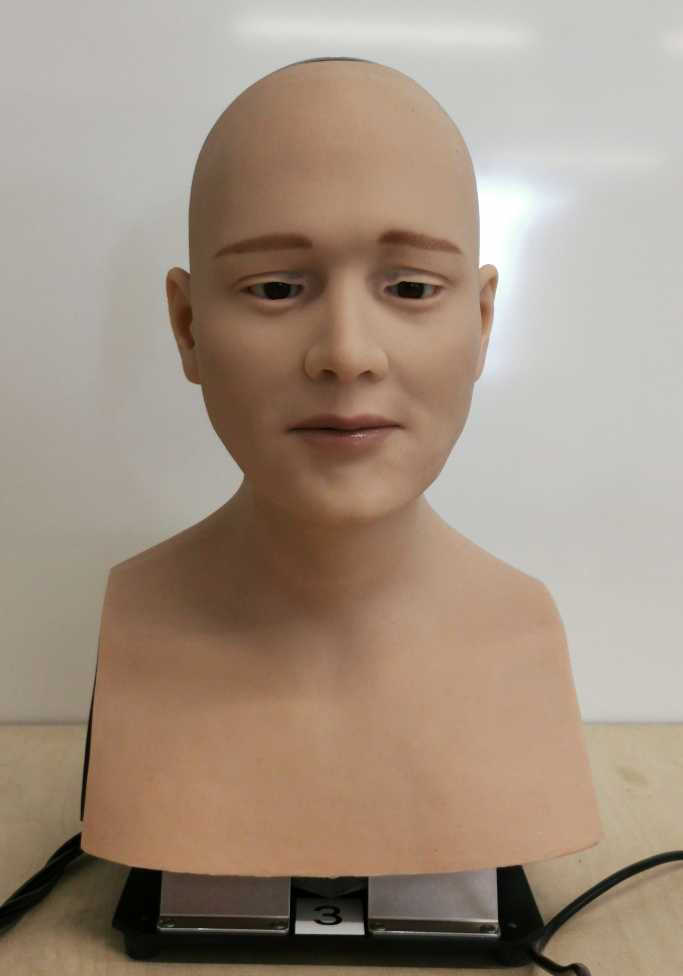}} &
\subfloat{\includegraphics[scale=0.1]{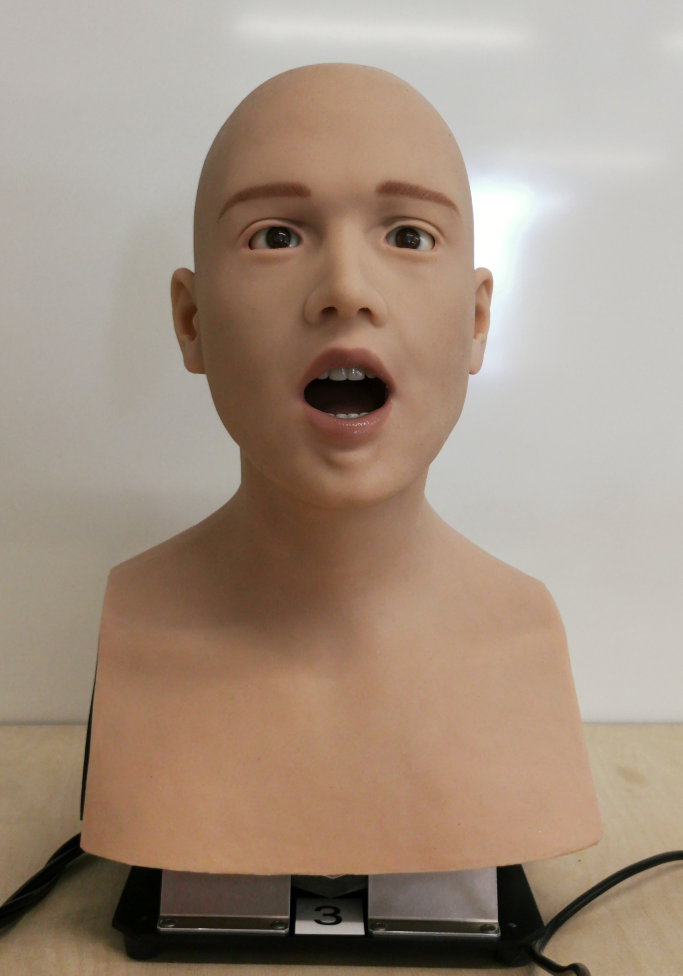}} &
\subfloat{\includegraphics[scale=0.1]{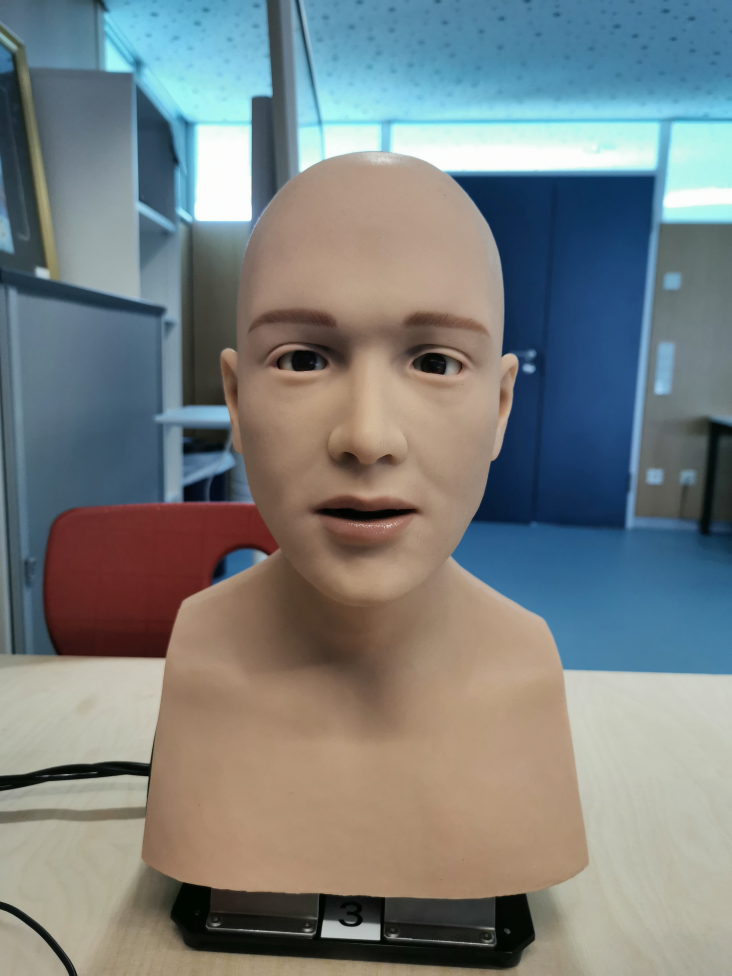}} &
\subfloat{\includegraphics[scale=0.1]{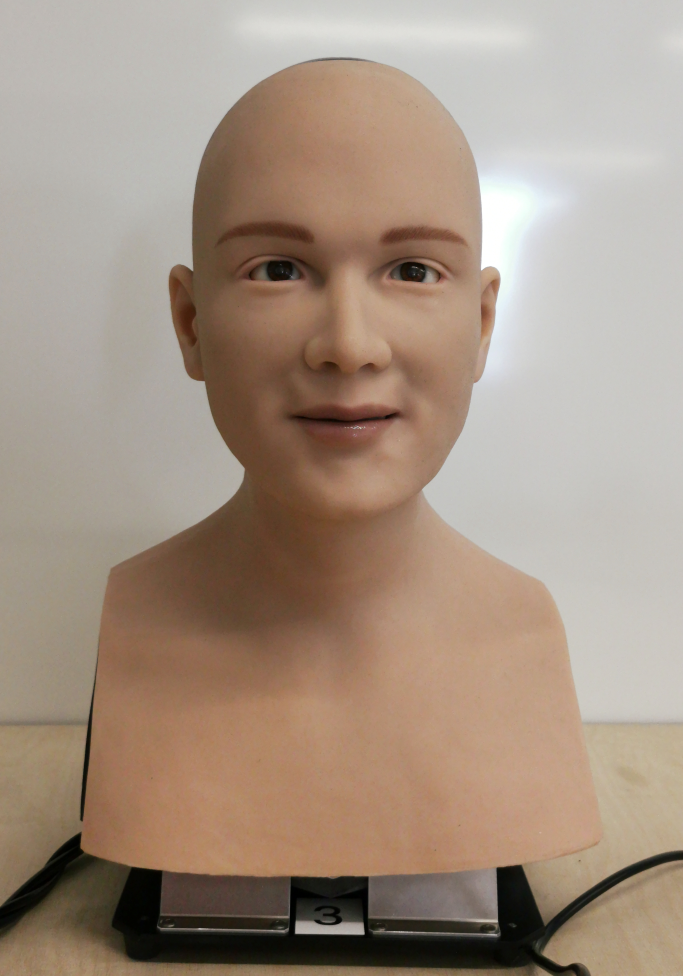}} \\
\subfloat{\includegraphics[scale=0.1]{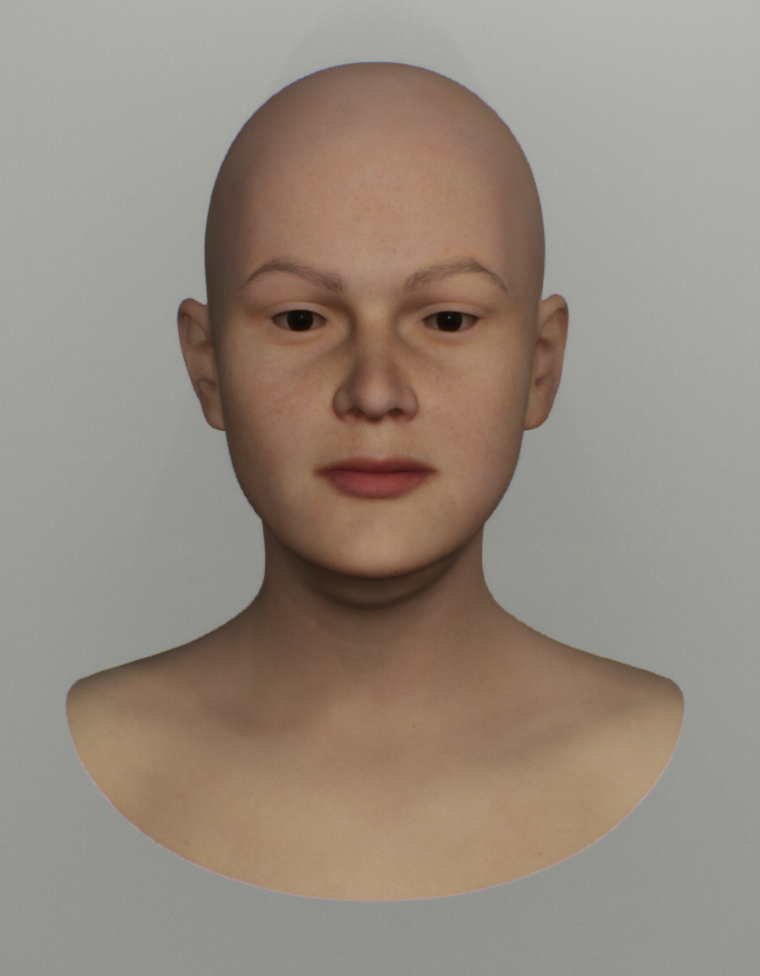}} &
\subfloat{\includegraphics[scale=0.1]{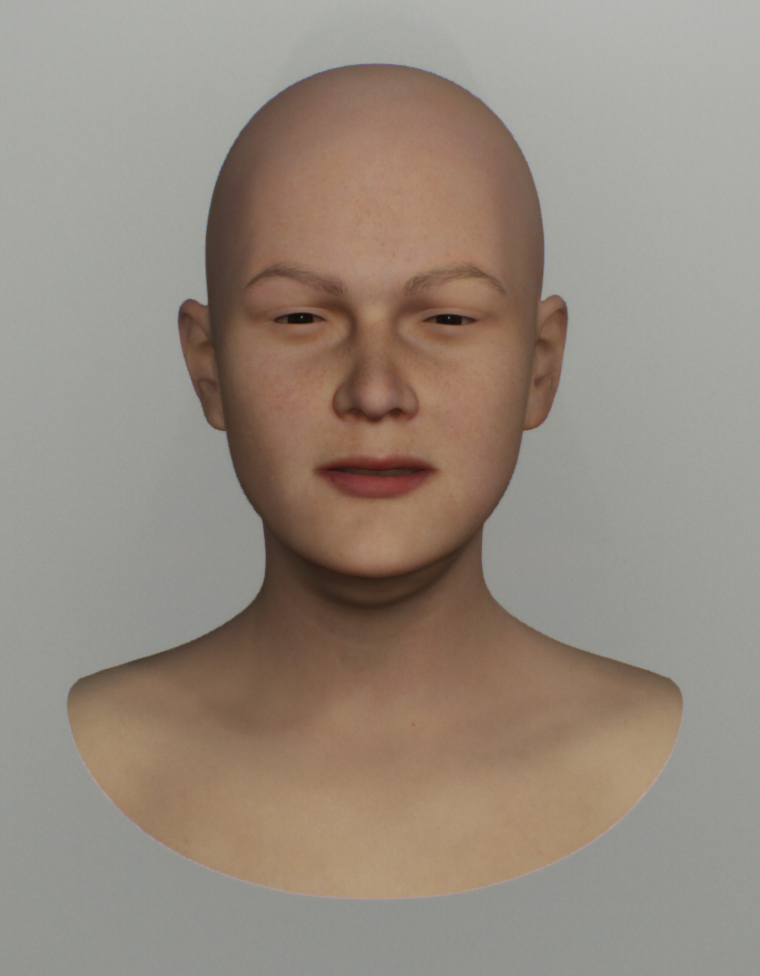}} &
\subfloat{\includegraphics[scale=0.1]{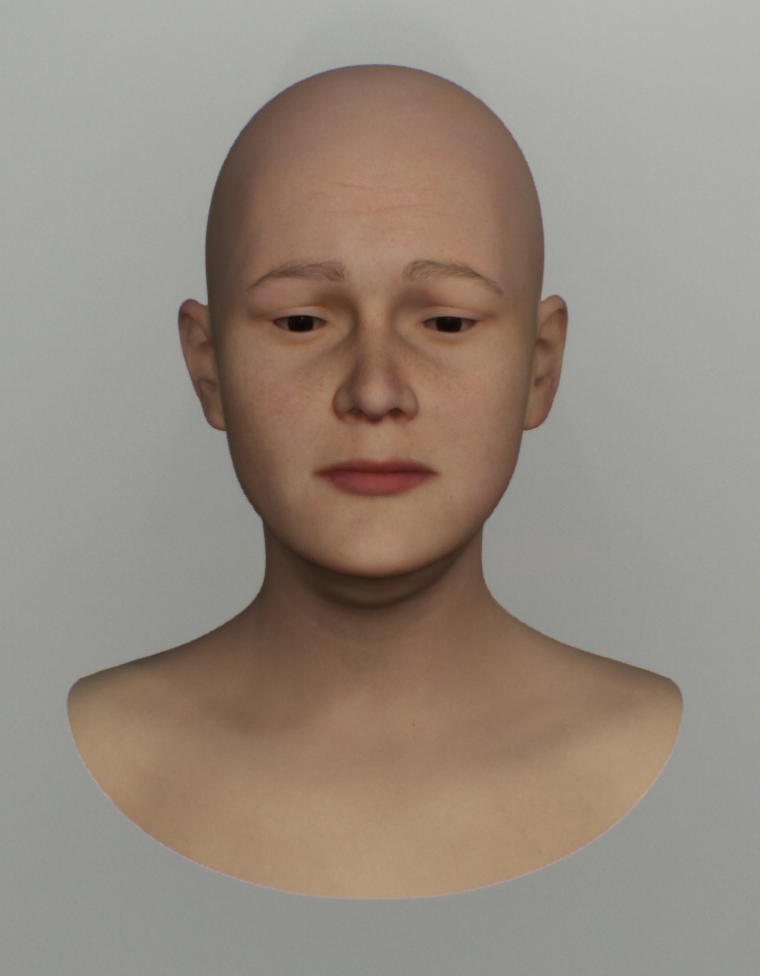}} &
\subfloat{\includegraphics[scale=0.1]{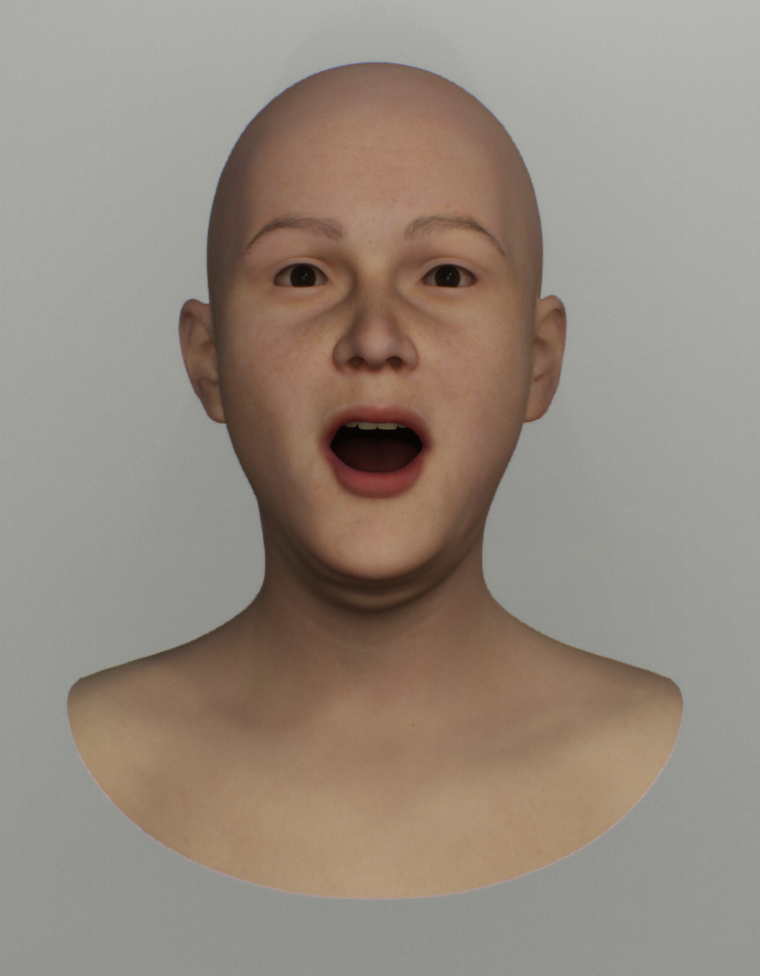}} &
\subfloat{\includegraphics[scale=0.1]{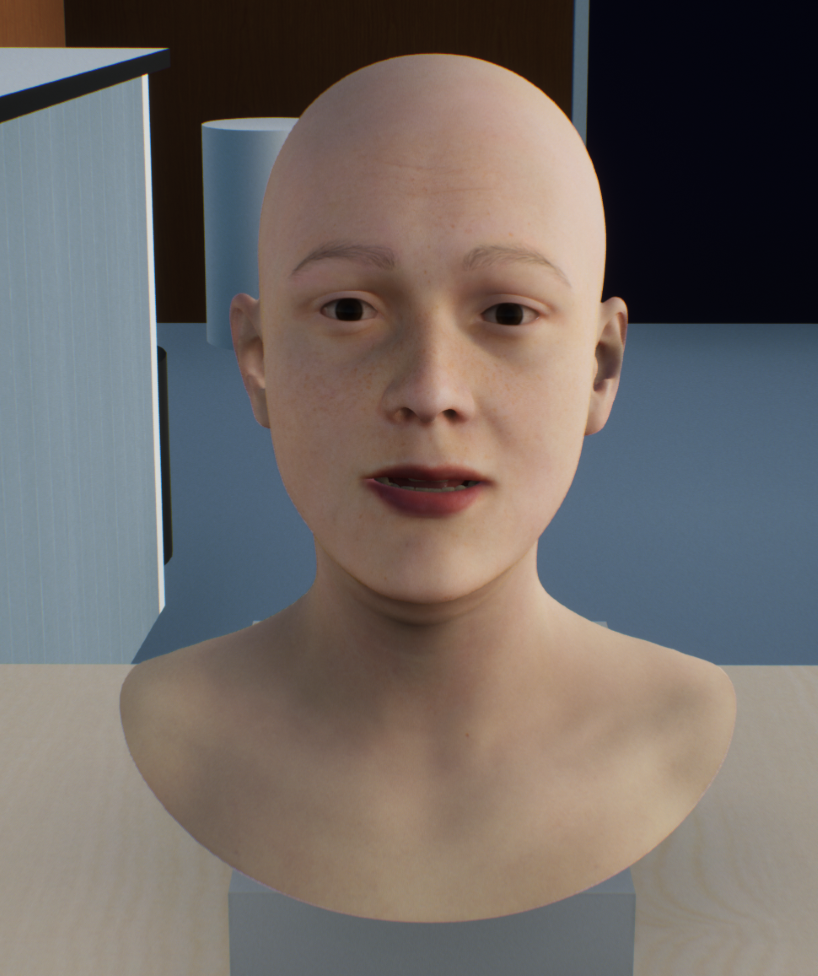}} &
\subfloat{\includegraphics[scale=0.1]{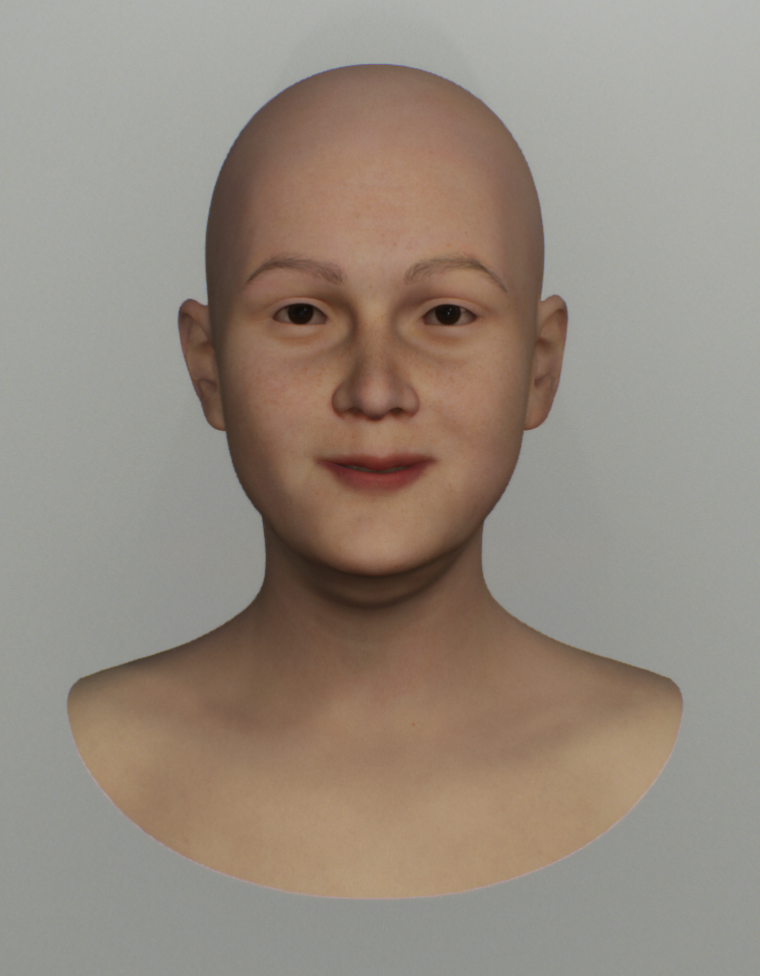}} \\
\subfloat{\includegraphics[scale=0.1]{00_Neutral_cropped.png}} &
\subfloat{\includegraphics[scale=0.1]{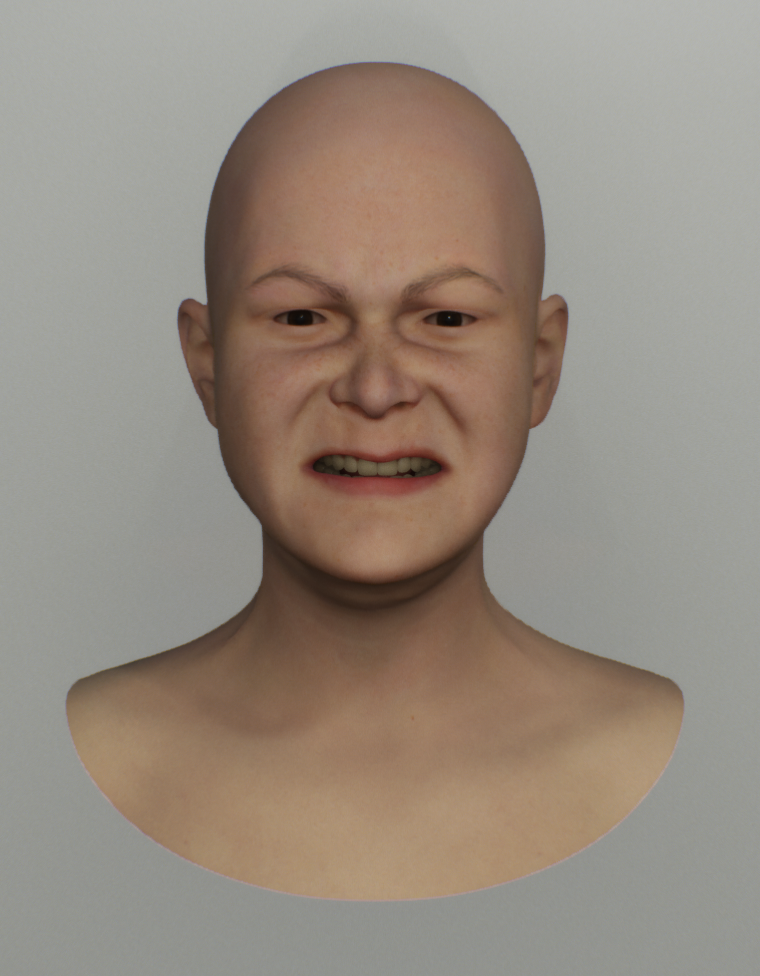}} &
\subfloat{\includegraphics[scale=0.1]{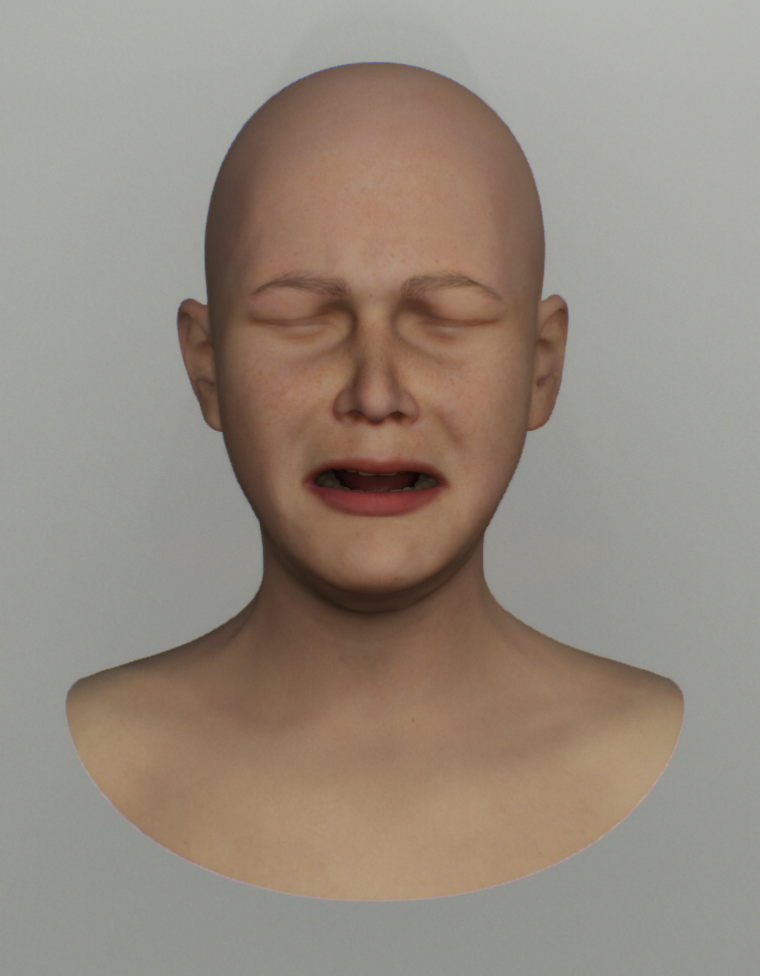}} &
\subfloat{\includegraphics[scale=0.1]{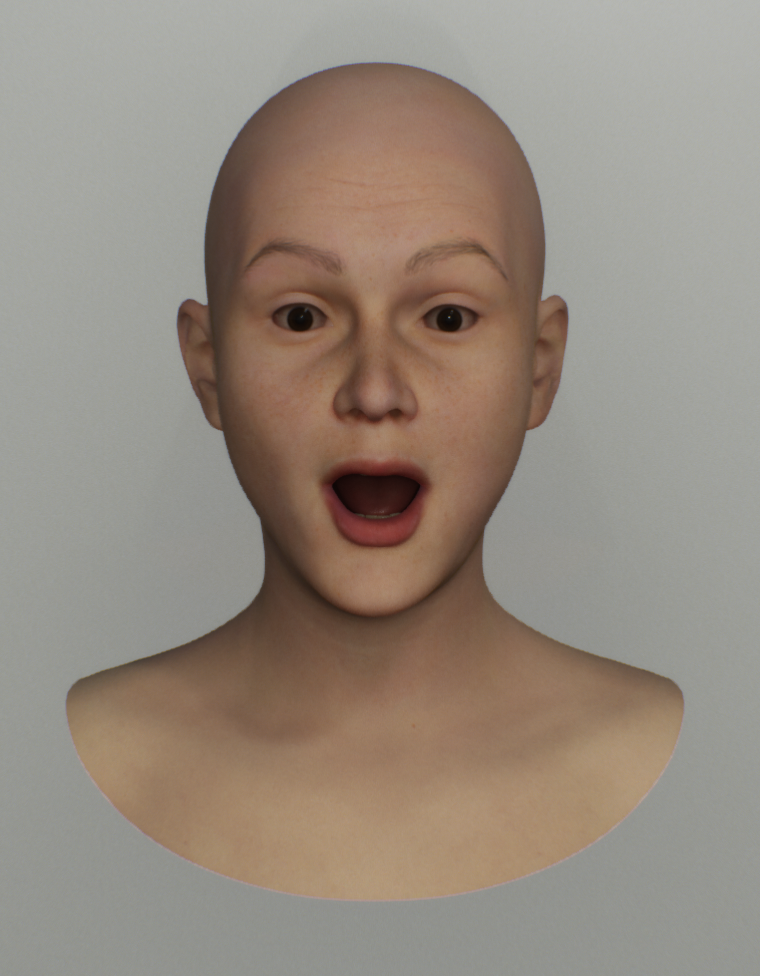}} &
\subfloat{\includegraphics[scale=0.1]{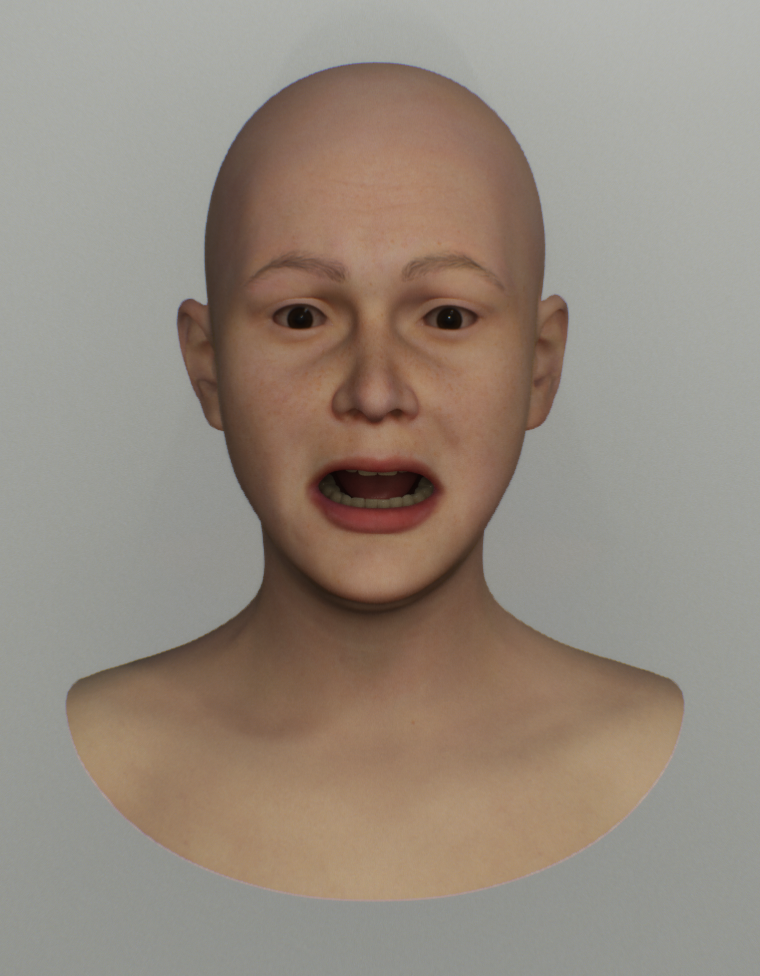}} &
\subfloat{\includegraphics[scale=0.1]{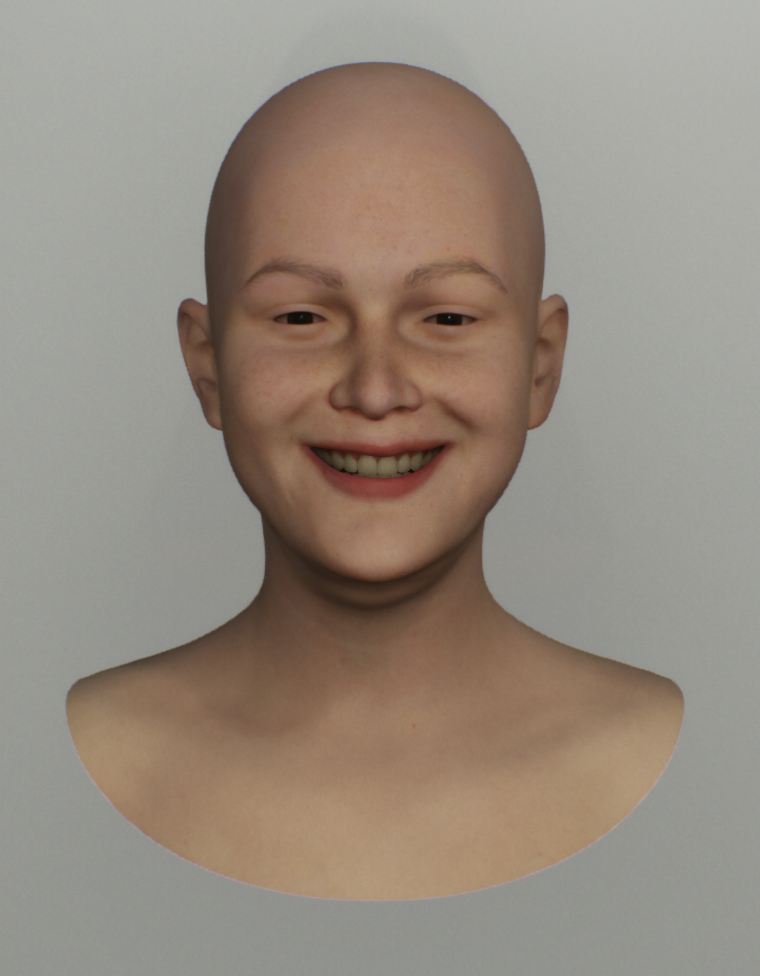}}
\label{fig3}
\end{tabular}
\caption{Emotions used in the main study, on the real robot head (top), recreated on the digital robot head (middle), and ``ideal" emotions taken from Unreal's Pose Library (bottom); From left to right: neutral, anger, sadness, surprise, fear (as changed after pre-study), and happiness.}
\label{fig-overview}
\end{figure*}

The basic emotions of anger, sadness, surprise, fear, and happiness (cf.~\cite{b11}) were shown both on the real and virtual robot head, with each emotional expression presented twice per condition, cp.~Fig.~\ref{fig-overview}.

\renewcommand{\arraystretch}{0}

To keep the emotion ratings consistent across all three conditions, participants were asked over a speaker using the synthetically generated words "anger," "sadness," "surprise," "fear," and "happiness." The participants were then to give a verbal rating on a scale from 0 (not at all) to 4 (completely) for each of the emotions just heard, which were always asked in this order for each facial expression. The procedure was based on \cite{b18}. The same five-point scale was presented in both, the real and virtual conditions, as a piece of (real or virtual) paper placed in front of the participants, see $S$ in Fig.~\ref{fig2}.

\subsection{Experimental setup}
\begin{figure}[htbp]
\centerline{\includegraphics[scale=0.7]{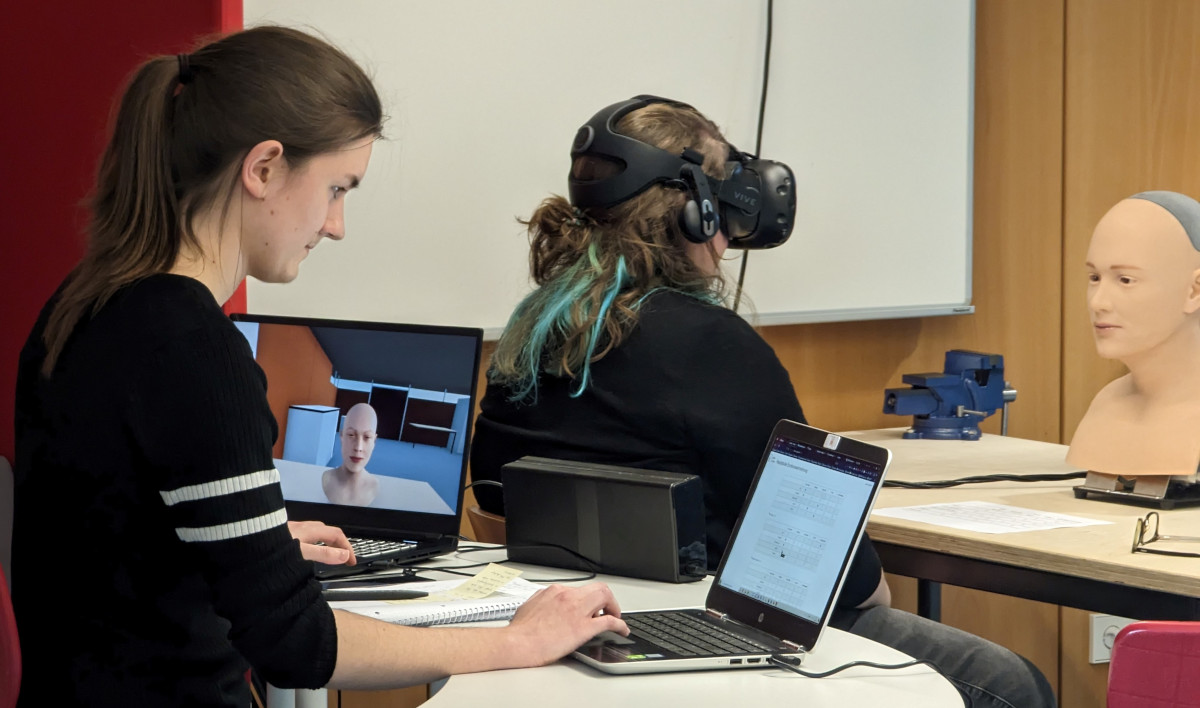}}
\caption{Experiment setup for the main experiment (side view); The experimenter controls the virtual head's facial expressions, left, and notes the participant's responses, right.}
\label{fig_sideview}
\end{figure}

\begin{figure}[htbp]
\centerline{\includegraphics[scale=0.4]{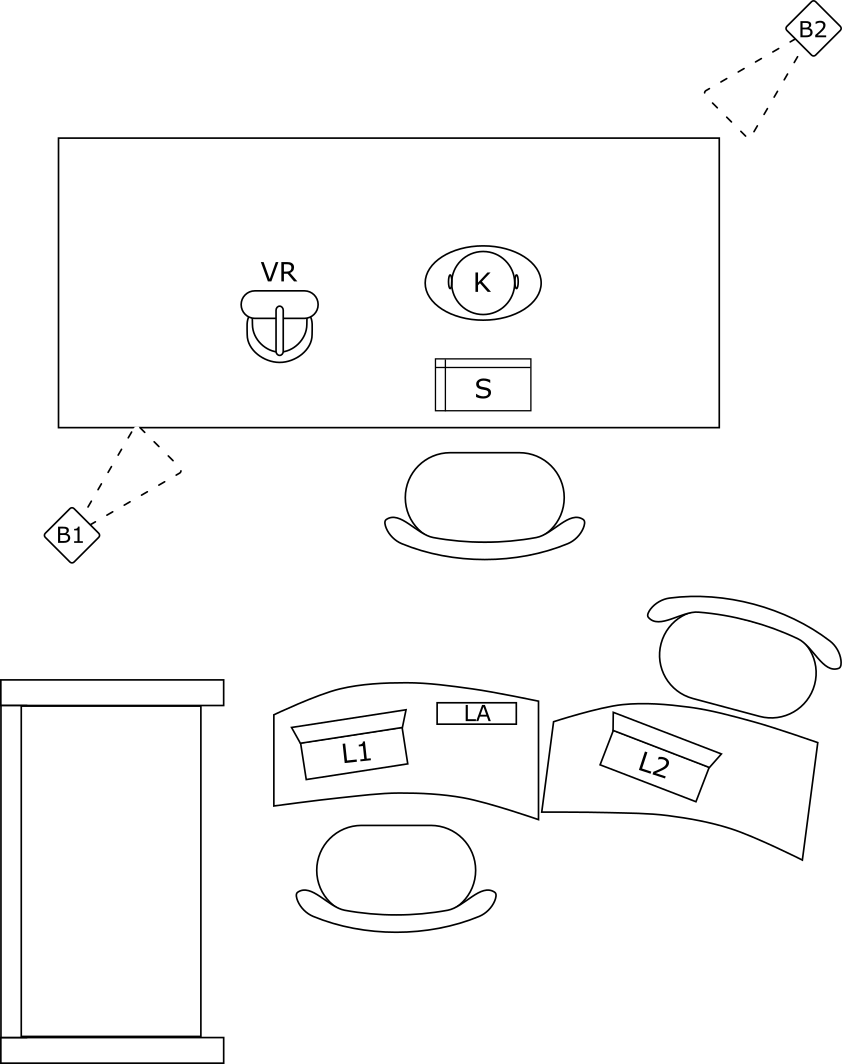}}
\caption{Experiment setup for the main experiment (top view); $L1$ main laptop for control, $L2$ laptop with questionnaire, $LA$ loudspeaker for playing asked emotions, $K$ robot head, $S$ scale, $VR$ VR goggles, $B1$ and $B2$ base stations.}
\label{fig2}
\end{figure}

The experiment took place from May 13 to May 18, 2023 at the Humanoid Lab of the Stuttgart Media University. 
The participants sat face-to-face with the physical robot head and were able to view the virtual robot head via the HTC Vive VR-headset. Using laptop L1 (cp.~Fig.~\ref{fig_sideview} and Fig.~\ref{fig2}), the real head was controlled via a GUI and the virtual head via the Unreal scene from an operator behind the participant. On a second laptop L2 the corresponding questionnaire was first filled out by the participant himself and for the emotion rating by the experimenter.

\subsection{Participants}
Thirty individuals participated in the main study, 16 of whom identified as female ($M = 24.19$, $SD = 4.85$) and 14 as male ($M = 25.29$, $SD = 4.61$). Recruitment was done via the university mailing list and among friends, targeting students and individuals with academic backgrounds. 

Six persons had participated in the pre-study, but since it was not resolved which emotion was assessed, this was not seen as an exclusion criterion. Only one person ($3.33\%$) had repeatedly worked with an android robot in the last six months, four others ($13.33\%$) rarely. For 12 people ($40\%$) VR was a new experience, 16 people ($53.33\%$) had rarely done anything with it. Two persons ($6.66\%$) had repeatedly dealt with it.

\subsection{Experimental procedure}
First, participants were shown the robot head and the evaluation scale. The oral evaluation procedure was explained to them. They were told that the robot head starts from the neutral position, moves into an expression, and stays in that expression until all scores for it have been given. 
In the questionnaire, participants were informed about the use and processing of their data with a privacy statement. Subsequently, participation had to answer questions regarding their participation in the pre-study, age, gender, experience with android robots, and VR in the last six months. 
Using a habituation condition, participants were supposed to internalize the scale and practice responding verbally. They were presented with an printed emoji and asked to rate it according to the scale. Any ambiguities could be clarified afterwards, and questions were also possible at any time during the conditions.

Each of the 30 participants was shown three conditions with 10 facial expressions each (two each of anger, sadness, surprise, fear, happiness). For each of the emotional expressions, they were asked to rate on a scale of 0-4 how strongly it corresponds to each of the five emotion labels. 

The emotional expressions were triggered in random but pre-defined order by the experimenter as described above and the ratings were queried and entered into the questionnaire. 
After all three conditions had been completed, the participants were asked whether there were any problems, and if so, this was noted. If desired, the background of the work was explained. Otherwise, thanks were given for participation and participants could pick some sweets.

\begin{table*}[t]
\caption{Results for the Shapiro-Wilk-Test, the first Friedman and the Dubin-Conover-Test}
\begin{center}
\begin{tabular}{|c|c|c|c|c|c|}
\hline
&\multicolumn{5}{|c|}{\textbf{analyzed emotion}} \\
\hline
&\textbf{\textit{anger}}& \textbf{\textit{sadness}}& \textbf{\textit{surprise}}& \textbf{\textit{fear}}& \textbf{\textit{happiness}} \\
\hline
Shapiro-Wilk & $p < 0.007$& $p < 0.002$& $p < 0.0017$& $p < 0.05^{\mathrm{*}}$& $p < 0.02$\\
\hline
Friedman E& $\chi^2 = 87.4; p < 0.001$ & $\chi^2 = 98.4; p < 0.001$& $\chi^2 = 84.1; p < 0.001$& $\chi^2 = 91.8; p < 0.001$& $\chi^2 = 93.0; p < 0.001$\\
\hline
Friedman VN& $\chi^2 = 79.9; p < 0.001$& $\chi^2 = 99.3; p < 0.001$& $\chi^2 = 75.7; p < 0.001$& $\chi^2 = 67.2; p < 0.001$& $\chi^2 = 94.0; p < 0.001$\\
\hline
Friedman V& $\chi^2 = 77.4; p < 0.001$& $\chi^2 = 99.1; p < 0.001$& $\chi^2 = 90.5; p < 0.001$& $\chi^2 = 84.6; p < 0.001$& $\chi^2 = 93.6; p < 0.001$\\
\hline
Durbin-Conover& $p < 0.001$& $p < 0.001$& $p < 0.001$& $p < 0.002^{\mathrm{**}}$& $p < 0.001$\\
\hline
\multicolumn{5}{l}{$^{\mathrm{*}}$Deviations for sadness E ($p = 0.05$), surprise VN ($p = 0.086$), fear E ($p = 0.054$), and anxiety V ($p = 0.057$)}\\
\multicolumn{5}{l}{$^{\mathrm{**}}$Deviations in the comparison between fear and sadness (condition E with $p = 0.860$, condition VN with $p = 0.699$)}\\
\multicolumn{5}{l}{\hspace{0.3cm}and fear and surprise (condition V with $p = 0.105$)}\\
\end{tabular}
\label{tab_friedman}
\end{center}
\end{table*}

\section{Results of the main study}
\label{sec-results}
The answers were analyzed by grouping them according to the emotions shown and forming a mean and a difference of the same ratings in each case. According to the results of a Shapiro-Wilk test, the Friedman test had to be used to test for significant differences between ratings. A second Friedman test was used for comparing the ratings between conditions. In the case of significant differences, a pairwise comparison with the Durbin-Conover post hoc test was also performed. The condition with the real robot head is abbreviated as "E", the condition with the virtual robot head with the replicated emotions is abbreviated as "VN", and the condition with the emotions adopted from Unreal is abbreviated as "V". Statistical analyses were performed using jamovi software and a significance level of $p = 0.05$ was used.

\subsection{Intended expression of anger}
Anger was most likely to be identified as anger in all three conditions. The value was highest for the "ideal" virtual representation V ($M = 3.00; SD = 1.05$). The replicated virtual representation VN achieved a mean value of 2.68 ($SD = 0.887$) for anger, closely followed by the representation on the real robot head E ($M = 2.68; SD = 0.965$). The representation on the real and replicated robot head was partially classified as surprise (real $M = 0.433; SD = 0.691$, replicated $M = 0.333; SD = 0.531$), in condition V more as sadness ($M = 0.750; SD = 0.716$).

In conditions V and VE, there were occasional differences of 3 or 4 points between the first and second ratings. All other differences, also for the other emotions, had a maximum delta of 2. The second Friedman test reveals that there are no significant differences between the three conditions for the rating of anger ($\chi^2 = 4.53; p = 0.104$).

\subsection{Intended expression of sadness}
Sadness has a very good detection rate, as the mean values for sadness are highest for E ($M = 3.52; SD = 0.533$), VN ($M = 3.38; SD = 0.739$), and V ($M = 3.92; SD = 0.265$), and the median is 3.5 for E and VN, and as high as 4 for V. Fear was most likely to be selected in second place, where the mean values are 0.85 ($SD = 0.811$) for E, 0.95 ($SD = 0.913$) for VN, and 1.15 ($SD = 1.07$) for V.

Only in the evaluation of fear there were isolated differences of 3 or 4 points, otherwise they deviated only rarely and always by a maximum of 2 points in the other evaluations. Here, the second Friedman test yields significant differences for the three conditions ($\chi^2 = 15.0; p < 0.001$), which is why a pairwise comparison could be performed. While the presentation of sadness and its evaluation show significant differences for V compared to E and VN ($p < 0.001$), there is no significant difference between E and VN ($p = 0.698$).

\subsection{Intended expression of surprise}
Surprise shows the best recognition rate of all emotions shown. The mean value is 3.93 ($SD = 0.173$) in condition V, 3.77 ($SD = 0.487$) in E, and 3.53 ($SD = 0.642$) in VN. The medians for surprise are all at the maximum value of 4. Fear was rated second highest in condition E ($M = 1.20; SD = 1.02$) and in condition VN ($M = 0.983; SD = 1.13$), and happiness in condition V ($M = 1.27; SD = 1.17$). The delta of the ratings differs only slightly in most cases. In the case of surprise, a difference of 3 occurred once each in conditions E and VN, and for fear a difference of 3 for condition V. Only in the case of happiness did a difference of 3 occur once for condition V and a difference of 4 for conditions VN and V.

Comparing the display of surprise across the three conditions with the second Friedman test, significant differences emerge ($\chi^2 = 11.8; p = 0.003$). However, according to the Durbin-Conover post hoc test, only the differences between the simulated virtual robot head and the real ($p = 0.017$) and "ideal" robot head ($p < 0.001$) are significant. No significant differences were found between the rating of surprise on the real robot head and the rating of surprise on the "ideal" virtual robot head ($p = 0.223$).

\subsection{Intended expression of fear}
Fear was recognized worst compared to the other emotion representations. The mean values are only 1.07 ($SD = 0.691$) for condition E and 1.18 ($SD = 0.933$) for condition VN. Only the mean of 2.82 ($SD = 0.771$) for Condition V is acceptable. For Condition E, ratings were most likely for surprise ($M = 2.50; SD = 0.881$) and sadness ($M = 1.10; SD = 0.759$). Likewise for condition VN (surprise $M = 2.13; SD = 1.06$, sadness $M = 1.27; SD = 0.971$). For condition V, the rating for surprise ($M = 2.45; SD = 1.12$) comes second. The ambiguous ratings of the representation of fear also become apparent in the deltas between the two ratings given. For anger, there is a difference of 3 in condition V. Differences of more than three points occurred for sadness, surprise as well as fear. Conditions E and V were affected for sadness and surprise, and only condition V was affected for fear.

The second Friedman test for anxiety, the comparison across conditions, reveals significant differences ($\chi^2 = 41.9; p < 0.001$). The Durbin-Conover post hoc test continues to show significant differences for ratings of anxiety between conditions E - V as well as conditions V - VN ($p < 0.001$). Only the ratings between conditions E - VN show no significant differences with $p = 0.4$. However, there were no significant differences for sadness ($\chi^2 = 0.804; p = 0.669$) and surprise ($\chi^2 = 0.804; p = 0.669$).

\subsection{Intended expression of happiness}
Happiness was not consistently rated with maximum values in all three conditions, but it was most likely to be rated as happiness. Condition V performed best here ($M = 3.47; SD = 0.692$), followed by Condition VN ($M = 3.02; SD = 0.7484$) and Condition E ($M = 2.95; SD = 0.834$). Other emotions tended to be less reported here, with a bit surprise for Condition E and VN (for Condition E with $M = 0.500; SD = 0.639$, for Condition VN with $M = 0.483; SD = 0.663$) and very little anger for Condition V ($M = 0.200; SD = 0.407$). There are no major differences in the deltas for happiness. For anger and surprise, there is a difference of 3 points once each in condition E, otherwise the ratings differ by a maximum of 2 points.

To investigate this further, a Friedman test was used to check whether there were differences in terms of ratings across conditions. Based on the previous results, only the rating for happiness was compared here. The differences were indeed significant ($\chi^2 = 13.4; p < 0.001$). The Durbin-Conover pairwise comparison specifically indicates that there are significant differences between conditions E and V ($p < 0.001$) and V and VN ($p = 0.002$). Only between condition E, the real robot head, and condition VN, the replicated virtual robot head, there are no significant differences with $p = 0.618$.

\section{Discussion and conclusion}
\label{sec-discussion}
The virtual recreation of an android robot head for comparison of emotion perception was presented. A virtual model was created using a 3D scanner and animated using the MetaHuman framework in Unreal Engine 5. The aim was to investigate whether the emotions displayed on the heads are generally perceived differently and whether more expressive emotional faces on the virtual robot head lead to better recognition rates.

Three conditions were designed, first, the emotions represented on the real head (E), second, the same emotions replicated on the virtual head (VN), and, third, the virtual head with "ideal" emotions that are not bound by the limitations of the physical hardware (V). The basic emotions anger, sadness, surprise, fear and happiness were chosen as emotional facial expressions and were validated in a pre-study.

Except for anger, significant differences across conditions were found for all emotions, supporting the first hypothesis. The assumption that the ``ideal" virtual head performs better due to more display options (hypothesis 2) was confirmed, since there were always significant differences between the conditions V and VN and the mean value was higher for V in each case. Therefore, the emotional expressions of the Unreal pose library can serve as an orientation for adjustments of the real robot head's emotional expressions. However, the assumption of the real robot head performing better than the replicated robot head cannot be confirmed (hypothesis 3). Except for the rating of surprise, for which the real robot head was rated better, no significant differences were found. This suggests that the type of presentation (real/virtual) does not always influence the perception of dynamic emotional expressions. 
The fourth hypothesis, namely that the "ideal" virtual robot head is rated better than the real robot head, can be confirmed for the emotions sadness, fear, and happiness. For surprise, no such significant differences in the ratings were found.


The finding that surprise, sadness, and happiness can be recognized well while fear seems more difficult to be expressible by facial expression alone is consistent with previous findings (cf.~\cite{b17, b28}). For fear, it should be noted that the confusions already occurred in the pre-study and the adaptation of the presentation did not lead to better recognition. The physical head might be limited by its hardware and movement capabilities, so any revision of the head should most likely focus on representation capabilities for fear. Additional modalities such as speech (high pitch) or sounds (gasping, shouting) could also support the portrayal of fear. 

\section*{Ethical Impact Statement}

All participants provided written consent to provide their data for the sole purpose of the empirical study. They were informed that the data will be kept securely according to the university's privacy rules and guidelines. As described above, we are aware that the results presented here are of limited generalizability, because of the participants' limited diversity. Also, a robot portraying emotional facial expressions might impact a future society in several, unforeseen ways and needs to be investigated as well.

\section*{Acknowledgment}

We thank Jens Hahn for the insightful advise during the project phase. 

\end{document}